\documentclass[reqno,12pt]{amsart}
\usepackage[utf8]{inputenc}
\usepackage[T1]{fontenc}
\usepackage{amsmath,amsfonts,amssymb,amsthm}
\usepackage{geometry}
\usepackage[normalem]{ulem}
\usepackage{tikz}
\usepackage{graphicx}
\graphicspath{{./images/}}
\usepackage[pdftex,
            colorlinks=true,
            linkcolor=blue,
            citecolor=red,
            pdfauthor={},
            pdftitle={},
            pdfcreator={pdflatex}]{hyperref}
\usepackage{verbatim}
\usepackage{color}

\usepackage[backend = biber, sorting = nyt, citestyle = numeric, bibstyle = numeric, maxnames = 50]{biblatex}
\usepackage{mathrsfs}
\definecolor{verde}{RGB}{20,150,100}

\newcommand{\R}{\mathbb R}

\numberwithin{equation}{section}
\theoremstyle{plain}

\theoremstyle{remark}

\title{Neural networks for spectral optimization}

\author[A. de Villeroché, B. Bogosel, S. Breuils, D. Bucur, J.O. Lachaud]{Alexis de Villeroché, Beniamin Bogosel, Stéphane Breuils, Dorin Bucur, Jacques-Olivier Lachaud}

\begin{document}
\begin{abstract}
Given a functional dependent on the spectrum of a differential operator, we address the problem of finding a domain which optimizes this functional. PDE solvers might be used to tackle this optimization. It is however computationally expensive. We propose two neural network models which learn the spectrum directly from the geometry of the domain and can be used to optimize the domain from one or more eigenvalues. We investigate two representations. The first encodes the domain through Fourier coefficients and a light MLP, which is efficient on star-shaped geometries, achieving a precision of 0.2\%. Through a rescaling of the coefficients the designed models satisfy the scaling law of the eigenvalues. Additionally, averaging the outputs of the trained surrogates over rotations and reflections induces invariance for these transformations. The second is a model that takes the landscape function, the indicator function and the gradient of the landscape function. A Gram-Schmidt process produces orthogonal eigenfunctions as output of the model along with the associated eigenvalues. The landscape model reaches 1\% mean relative error on the first ten eigenvalues, compared with 4\% for an FNO model. Replacing the landscape by an SDF worsened both prediction and optimization errors. The trained model also generalizes from synthetic shapes to domains given as classical image dataset. The resulting surrogates of both approaches recover classical spectral optima such as the disk for the first eigenvalue or the conjectured minima of higher eigenvalues. This confirms that our models produce accurate differentiable estimates of eigenvalues, which can be used in shape optimization problems involving spectral quantities.
\end{abstract}
\maketitle

\section{Introduction} \label{sec:intro}
Spectral geometry, which explores the interplay between the geometric properties of a domain and the spectrum of the differential operators defined on it, is a fundamental research field in mathematics. Its questions, one of the most famous being "Why are all drums round?", find applications in numerous domains ranging from physics and biology to computer science. 
A key problem in this field is shape optimization, which corresponds to the search for geometric shapes that optimize one or more eigenvalues of the Laplace operator. 

Understanding the spectrum of the Laplacian leverages many applications in geometry processing (denoising, shape matching, shape retrieval). Computing this spectrum with finite elements is accurate, but each new shape requires a new computationally expensive solve. Neural surrogates aim to replace that by a single and fast forward pass. However their accuracy depends on how the domain is represented.

One of the major state-of-the-art representation is the \textbf{Signed Distance Function}, $SDF_\Omega$, with $\Omega$ the domain.
Briefly, the Signed Distance Function is a function  $SDF_\Omega(\mathbf{x})$ that gives the shortest distance from a point $\mathbf{x}$ to boundary $ \partial \Omega$ of the shape, with the sign indicating whether the point is inside or outside. It is used in recent works in geometry processing and computer vision~\cite{ichimaru2025neural}. A major issue of this representation is its topological sensitivity. To illustrate this, let us consider a domain with an internal hole of radius $\epsilon$. As $\epsilon \to 0$ (capacity vanishing perturbation of radius $\epsilon$), the Dirichlet spectrum converges continuously to that of the domain without hole. However the $SDF$ will keep its zero value near the hole. 

Current state-of-the-art approaches for predicting spectral properties also often rely on \textbf{Fourier Neural Operators} (FNO)~\cite{li2020fourier} or \textbf{DeepONets}~\cite{lu2021learning}.

\textbf{DeepONets}: DeepONets are based on the universal approximation theorem for operators. The architecture consists of two sub-networks: a Branch net, which encodes the input function (e.g.,the indicator function of the domain $\Omega$), and a trunk net, which encodes the coordinates $x$ where the output function is evaluated. The operator is represented as a dot product between the outputs of these two networks.

\textbf{Fourier Neural Operators (FNO)} on the other hand perform spectral convolutions in the frequency domain. The kernel integral operator is computed by applying a Fast Fourier Transform (FFT) to the input, multiplying by a learned weight matrix in the latent frequency space, and then applying an Inverse FFT. 
Since this approach relies on global frequency transformations, FNOs often struggle to capture sharp, localized geometric features or complex boundary conditions (Dirichlet/Neumann). FNOs often suffer from artifacts near the edges and struggle to accurately resolve high-frequency geometric features that are crucial for the upper part of the Laplacian spectrum.

Finally, current approaches use \textbf{Variational Physics-Informed Neural Networks (V-PINNs)} for Eigenproblems. While traditional PINNs solve for the eigenfunction $u(x)$ given a shape~\cite{jin2021nsfnets, lu2021deepxde}, recent Variational PINNs~\cite{kharazmi2019variational} use the Rayleigh Quotient:
\begin{equation}
    \mathcal{L}_{Rayleigh} =\frac{\int_{\Omega} |\nabla u|^2dx}{\int_{\Omega} |u|^2dx}
\end{equation}
as a loss function. By minimizing this functional, the model can learn the eigenvalues $\lambda_k$ as the stationary values of the optimization process. However, V-PINNs are limited to finding the first few eigenvalues and require a new optimization for every new shape.
 
\subsection{Our contributions}
In this paper, we propose differentiable surrogates for the first Dirichlet eigenvalues and eigenfunctions, with the explicit goal of shape optimization. We compare two representations. The first uses Fourier coefficients to represent the domain and has a light MLP structure that exploits translation and rotation invariance of specific geometries, its purpose being to quickly generate prototypes of optimal shapes for general functions. The second approach is a topology-aware encoder-decoder that takes the landscape function $w_\Omega$~\cite{AFST_2025_6_34_2_315_0}, the indicator $\mathbf{1}_\Omega$ and $\nabla w_\Omega$, and predicts $(\lambda_k,\phi_k)$ with a differentiable Gram-Schmidt step. We show that the landscape representation outperforms an SDF encoding and other state-of-the-art models, recovers classical optimization outcomes such as the disk for $\lambda_1$, and applies to domains given as images.

\section{Background and related work} \label{sec:Background}
On every bounded open set  $\Omega \subset \mathbb R^d $ one considers the Laplace operator with Dirichlet boundary conditions. The spectrum of the operators consists on a sequence of eigenvalues
$$0 <\lambda_1(\Omega) \le \lambda_2(\Omega)\le \cdots \to +\infty,$$
multiplicity being counted. These eigenvalues are the critical points of the Rayleigh quotient, namely 
$$\lambda_k(\Omega)= \min_{E \in \mathcal S _k} \max _{u\in E\setminus \{0\}} \frac{\int_{\Omega} |\nabla u|^2dx}{\int_{\Omega} |u|^2dx},$$
where $\mathcal S _k$ denotes the family of subspaces of dimension $k$ in the Sobolev space $H^1_0(\Omega)$. If the boundary of $\Omega$ is smooth, there exists a function $u_k \in H^1_0(\Omega)\setminus \{0\}$ such that
$$-\Delta u_k = \lambda_k(\Omega) u_k \mbox{ in } \Omega, \; u_k =0 \mbox{ on } \partial \Omega.$$
If the boundary of $\Omega$ is nonsmooth the equation above is satisfied in a weak, variational, sense.

In spectral geometry, mathematicians are interested in shape optimization problems which formally may be written as
$$\min _{\Omega \in\mathcal U_{ad}}F(\lambda_1 (\Omega), \dots, \lambda_k(\Omega)),$$
where $\mathcal U_{ad}$ is the family of admissible open sets and $F: \mathbb R^k\to \mathbb R$. A typical class $\mathcal U_{ad}$ of interest consists of all open subsets of $\mathbb R^d$ of prescribed measure.  The simplest such problem, for $F = \lambda_1 (\Omega)$, has a long history starting with the conjecture by lord Rayleigh from 1877 claiming that among all membranes with the same area the circular membrane (disk) has the lowest fundamental frequency (this encodes the circular shape of all drums). The conjecture was proved only in 1923 by Faber and Krahn, but opened the way to a series of deep conjectures relating geometry to the spectrum. Nowadays, these problems are studied by the spectral geometry community. While most of the questions are of pure mathematical nature, the theory of shape optimization initially developed by the engineering community provides effective tools to support or disprove these conjectures. However, running a shape optimization algorithm requires fine knowledges from PDEs, calculus of variations, optimization and scientific computing, making such procedures inaccessible to the most of mathematicians working in spectral geometry.

Several mathematical methods are available to approximate the eigenvalues of a given set $\Omega$: finite elements, finite differences, fundamental solutions, etc. and several shape optimization algorithms can be used (shape derivative, topological derivative, relaxation, level sets, etc). For each specific $F$, one has  to readapt the algorithms taking into account its specific features. The purpose of this paper is to propose a unified, simple method, based on neural networks which contains the full sequence (computations of eigenvalues - shape derivatives - shape optimization) and applies to a general $F$. 

A key question with a deep impact on the robustness of the computation, but also having its own mathematical interest, is the following: does a small geometric perturbation of $\Omega$ may have a strong impact on the variation of the spectrum? What is the right tool to capture the fine properties of the geometric dependence of the spectrum? Learning the spectrum from geometry and making this procedure stable, relies precisely on this analysis.

A first intuition is to take the characteristic function as an input for learning the spectrum. This is mathematically inconsistent with the constraint in terms of volume, since characteristic functions which differ up to a set of Lebesgue measure $0$ have the same $L^1$-norm. While the $L^1$ norms is stable for small variation of the characteristic function, the spectrum is very sensitive. A second possible input is the signed distance function to the boundary. This choice is again problematic, since a point in the middle of a disc would dramatically change the distance function, without any influence on the spectrum (points have $0$ capacity in dimension $2$). 

The choice we make is to use as main input the torsion (or landscape) function. For every $\Omega$ the landscape function $w_\Omega$ solves (in a weak sense if the boundary is not smooth, $w_\Omega \in H^1_0(\Omega)$)
$$-\Delta w_\Omega=1 \mbox{ in } \Omega, w_\Omega=0 \mbox{ on } \partial \Omega.$$
It has been mathematically proved that the landscape function controls the spectrum, precisely if $\Omega_1\subset \Omega_2$ then
$$|\lambda_k(\Omega_1)-\lambda_k(\Omega_2)|\le C\int w_{\Omega_2}-w_{\Omega_1} dx,$$
with a constant $C$ which is controlled.

Equipped with this representation, several recent works in operator learning relate to our method. \cite{rowan2025solving} exploited the Rayleigh quotient inside a neural architecture to compute eigenvalues, but, like the V-PINN approach discussed in the introduction, their network is re-optimized for every new shape. \cite{li2026finite} proposed a hybrid finite-element/neural framework (FEENet) that still requires a mesh of the domain, whereas the torsion-based encoding only needs an indicator function. \cite{costabal2024delta} and \cite{williamson2025neural} address related PDE-on-complex-geometry and geometry-processing problems, but do not target the Dirichlet spectrum or shape optimization. On the shape-optimization problem, \cite{belieres2025volume} used a variational neural network to perform volume-preserving optimization of the Dirichlet energy, which is methodologically close to our differentiable pipeline but targets a different (non-spectral) functional. Finally, the SDF representation discussed above is based on the representation presentated by \cite{park2019deepsdf} for shape reconstruction. Overall, unlike these approaches, our two method predicts the first $K=10$ eigenvalues in a single forward pass, for arbitrary domains given as a raster image, and is differentiable.

\section{Fourier coefficients based model} \label{sec:fourier}

\subsection{Approach}\label{sec:appFourier}
When performing numerical shape optimization in the plane, one of the most natural class of admissible shapes is the class of bounded, star shaped, open sets. These shapes can be described in polar coordinates by 
$$ \Omega = \{ (\phi, \rho) \in [0,2\pi)\times \R^+ ~|~  \rho < \rho_\Omega(\phi) \}, $$
where the function $\rho_\Omega$ is a generic positive, continuous, $2\pi$-periodic function on $\R$. As such, it can be expressed as a Fourier series, 
$$ \rho_\Omega (\phi) = a_0 + \sum_{i\ge 1} (a_i\cos(i\phi) + b_i \sin(i\phi)), $$
the set $\Omega$ being uniquely defined by the sequences of coefficients $(a_i)_{i\ge 0}$ and $(b_i)_{i\ge1}$. It is therefore natural to focus on sets defined by a fixed finite number of coefficients $2N+1$. In this study we work in the class of shapes 
$$ \mathcal{U}_{15}  = \Big \{ \Omega \subset \R^2 ~\Big|~ \text{ star shaped},\ \rho_\Omega (\phi) = a_0 + \sum_{i = 1}^{15} (a_i\cos(i\phi) + b_i \sin(i\phi))\Big\}. $$
A strong quality of this class is that even though it is of finite dimension each shape has an exact analytic formulation and is independent of any discretization used to numerically compute the eigenvalues. Additionally, computing the area of a shape $\Omega$ in the class $\mathcal U_{15}$ is explicit, indeed 
\begin{equation}\label{eq:AreaFormula}
|\Omega| = \pi a_0^2 + \frac\pi2 \sum_{i = 1}^{15} (a_i^2 + b_i^2).
\end{equation}
Yet, there is no trivial method to determine that a given truncated Fourier series stays positive on the whole interval $[0,2\pi)$. To create the database needed to train our models, we add a decay assumption of the coefficients $(a_i)_{0\le i\le 15}$ and $(b_i)_{1\le i \le 15}$ to ensure the positivity of the border function, namely,
\begin{equation}\label{eq:CoeffDecay}
\forall 1\le i\le 15,\qquad (a_i^2 + b_i^2)^\frac12 \le 0.75\, \frac{a_0}{i^2}.
\end{equation}

\begin{figure}[ht]
\centering
\includegraphics[width=0.135\linewidth]{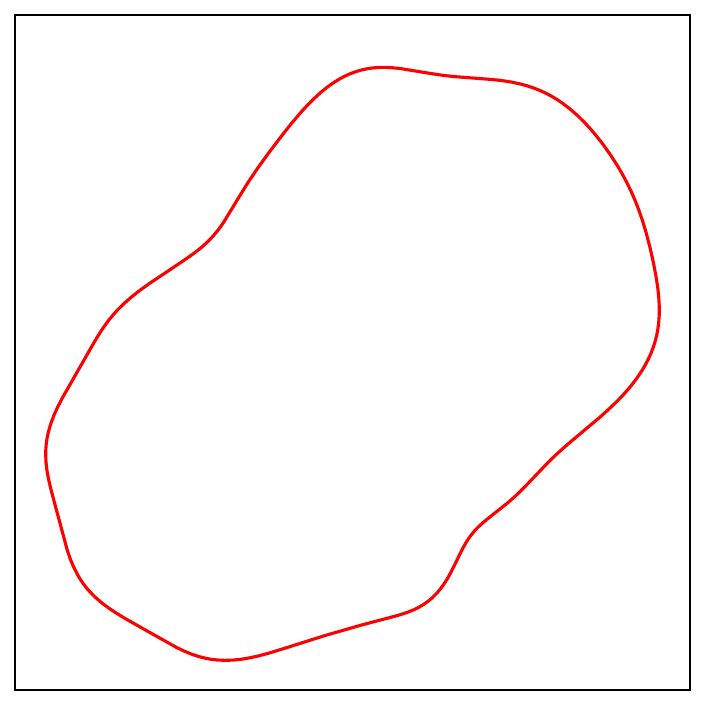}
\includegraphics[width=0.135\linewidth]{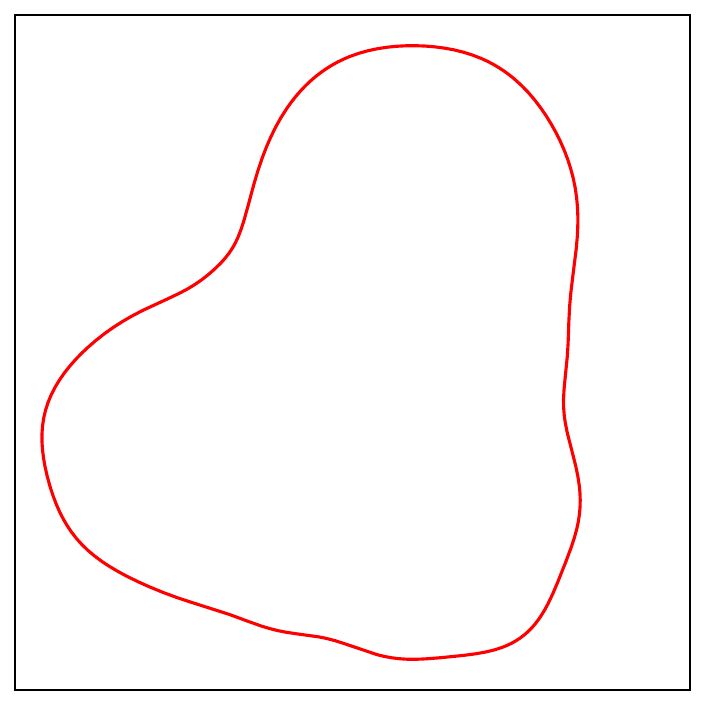}
\includegraphics[width=0.135\linewidth]{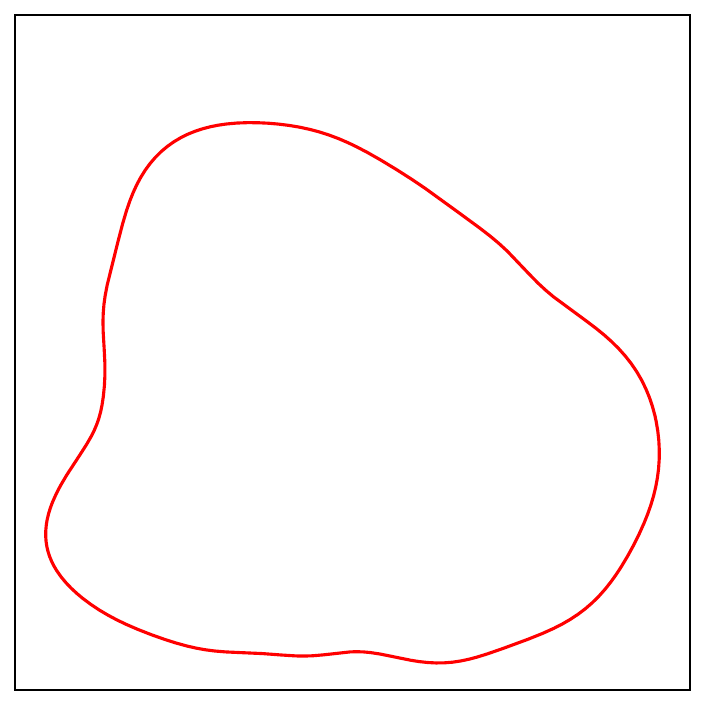}
\includegraphics[width=0.135\linewidth]{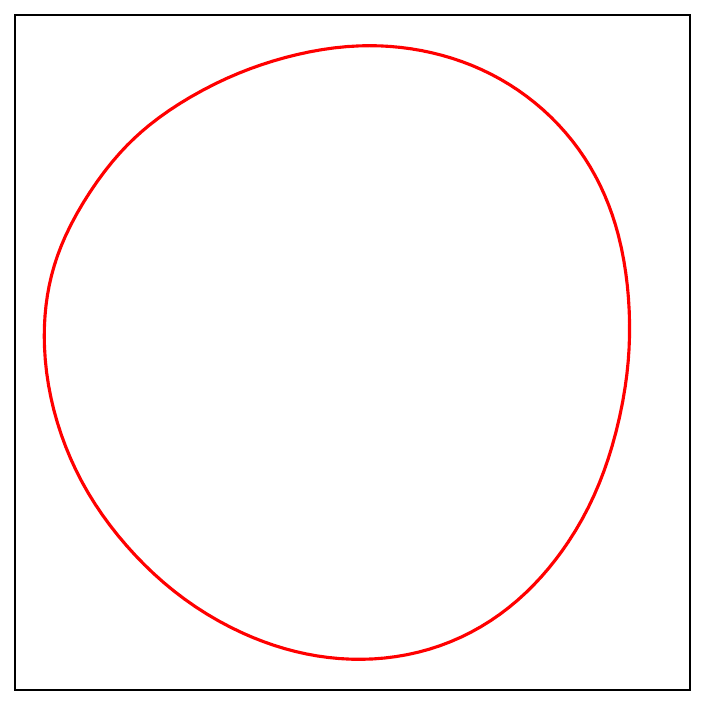}
\includegraphics[width=0.135\linewidth]{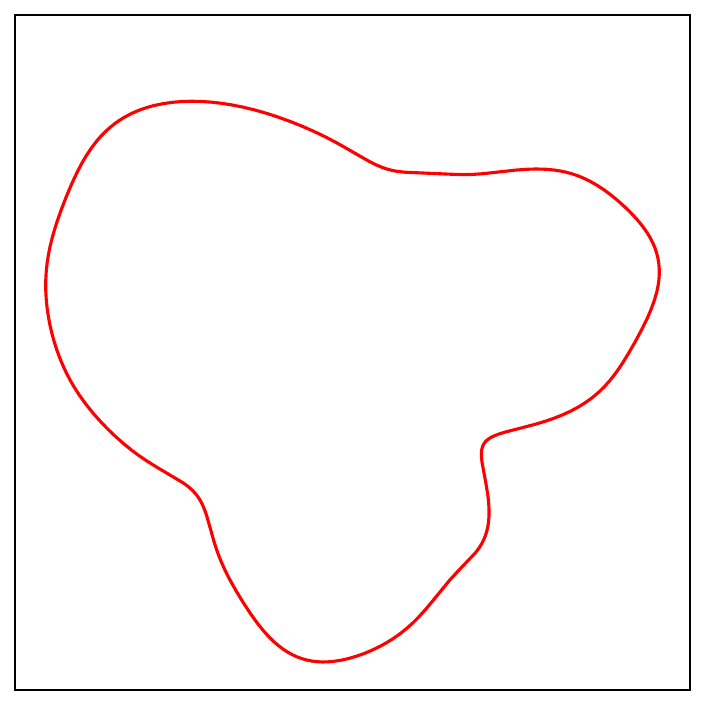}
\includegraphics[width=0.135\linewidth]{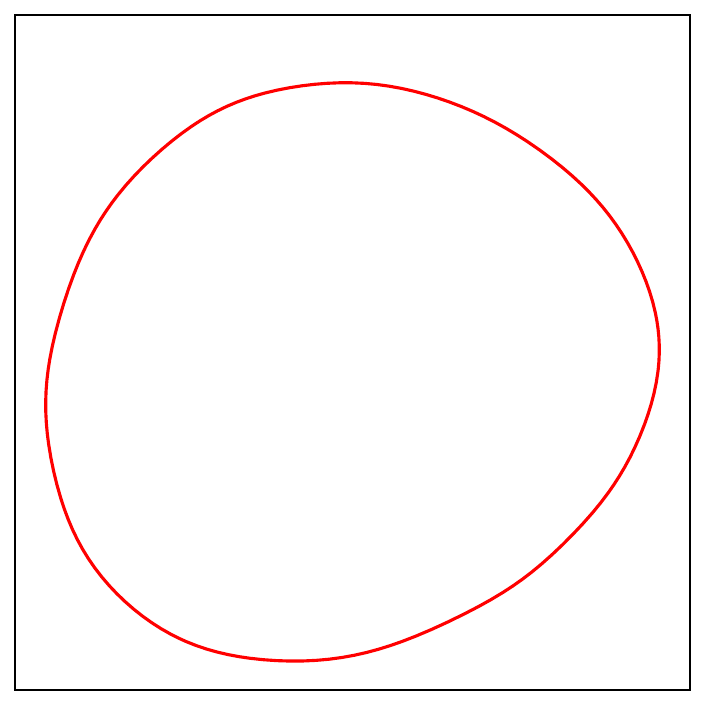}
\includegraphics[width=0.135\linewidth]{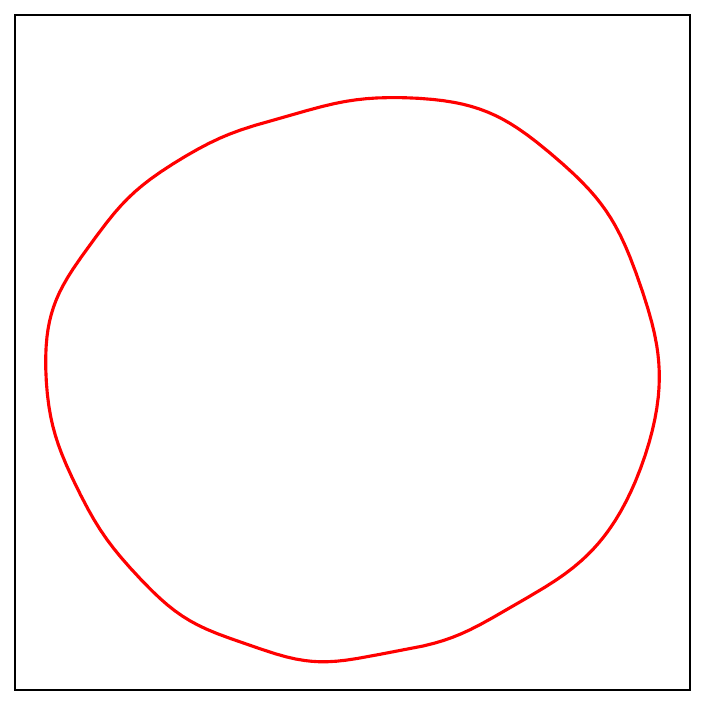}
\caption{Sample from the Fourier database.}
\label{fig:FourierShapes}
\end{figure}

We generate a database of $20 000$ entries (see Figure~\ref{fig:FourierShapes}) satisfying this decay assumption and computing the first $10$ Dirichlet eigenvalues of each entry using P2 elements and a finite element PDE solver written in FreeFEM~\cite{hecht2012new}.

Our aim is to build a surrogate to the PDE solver that computes the list of eigenvalues $L(\Omega) = \{\lambda_1(\Omega), \dots,\lambda_{10}(\Omega)\}$ given the parameters $(a_i)_{0\le i\le 15}$ and $(b_i)_{1\le i\le15}$ that we couple with the symbolic differentiation library SymPy to perform the optimization of a generic functional $F(\lambda_1, ...,\lambda_{10})$. The surrogate needs not only to be accurate but also to be sufficiently stable to perform optimization from its partial derivatives. Additionally, it needs to satisfy some qualitative properties of the eigenvalues: their scaling law and their invariance by rotations and reflections. Note that we do not try to impose invariance under translations of the shape, indeed for a given shape $\Omega$ in $\mathcal U_{15}$ the translated shape $\Omega' = \Omega + x$ is in general not in $\mathcal U_{15}$.

Concerning the scaling law, for each $k$ and each $\Omega\subset \R^2$ the $k$-th eigenvalue satisfies for any $t >0$ the relation $\lambda_k(t\Omega) = t^{-2}\lambda_k(\Omega)$ which implies in particular that the normalized eigenvalue $j_k(\Omega) = |\Omega| \lambda_k(\Omega)$ is invariant for dilations of $\Omega$. Noting that the area of a given shape $\Omega$ can be explicitly computed from its coefficients by \eqref{eq:AreaFormula}, we decide to train models that do not learn directly the list $L(\Omega)$ but rather the associated list of normalized eigenvalues $J(\Omega)$. Because of this normalization, we also notice that it is not needed to keep track of the first coefficient to compute $J$ as it is uniquely determined by the reduced set of coefficients $q = (a_1/a_0, ..., a_{15}/a_0, b_1/a_0, ..., b_{15}/a_0)$. We then design the surrogate to compute $J$ given the data of a vector $q \in \R^{30}$, and it satisfies the scaling law directly from its structure. 

To increase the stability of the derivatives of the surrogate we aggregate the results given by three independently trained MLPs. Each MLP has the same structure : $30 \to 192 \to 192 \to 192 \to 10$. Both the width and the activation function have been tested for performance and SiLU has been retained (see Section~\ref{sec:expFourier}).  The MLP is trained to compute for a given $q$ the quantity $\log J$ (the $\log$ is applied term by term on the list $J$) and we denote by $\log J(q)^{(m)}$ the output of a network $(m)$.  

To account for the invariance by reflection and rotations we average the output of the network $(m)$ over $12$ fixed rotations and their associated reflections for any given shape $\Omega$. Doing so, we compute the function

$$ \overline {\log J(q)^{(m)}} = \frac1{\#\mathcal T} \sum_{\tau \in \mathcal T} \log J(\tau(q))^{(m)}, $$

where $\mathcal T$ denotes the set of transformations corresponding to the reflections and rotations and $\#\mathcal T$ its cardinal. This averaged function, $q\mapsto \overline {\log J(q)^{(m)}}$, is by design invariant by reflections of the shape and by the twelve rotations. This induces the needed reflection invariance and a close to rotational invariance to the final surrogate without the necessity of learning it in the weights of the networks. 

Finally, we aggregate the three networks computing the following average 
$$ \tilde J(q) = \frac 13 \sum_{(m)} \exp(\overline {\log J(q)^{(m)}}). $$

\subsection{Experiments}\label{sec:expFourier}
To test our surrogate, we use two different types of validation. We first validate the model against test data unused during the training process for its accuracy. We then validate the model through optimization experiments, we check that it produces the expected sets and values compared to the shape optimization literature. As comparison we use the values obtained in \cite{AF12} in a similar class of sets but with a more traditional shape gradient based optimization process initialized with a genetic algorithm. For reference we give in Table~\ref{tab:ValAF12} the optimal values they obtained for $\{\lambda_5, ..., \lambda_{10}\}$.

\begin{table}[ht]
\caption{Optimal values for $\{\lambda_5, \dots,\lambda_{10}\}$ computed by Antunes and Freitas}
\label{tab:ValAF12}
\centering
\begin{tabular}{cccccc}
$\lambda_5 = 78.20$ & $\lambda_6 = 88.52$ & $\lambda_7 = 106.14$ & $\lambda_8 = 118.92$ & $\lambda_9 = 132.68$ & $\lambda_{10} = 142.72$ \\
\end{tabular}
\end{table}

We first perform tests to determine the best structure for the MLP. We test different activation functions and widths for the hidden layers. We test for the activation function in \{ReLU, SiLU, Tanh\} and the width in $\{90, 192, 300\}$ for networks computing the list of the first $10$ rescaled Dirichlet eigenvalues. For each couple (activation, width), we independently train three MLP networks of structure : $30\to \text{witdh} \to \text{witdh} \to \text{witdh} \to 10$ for $1200$ epoch and the best epoch is kept. We measure the total time needed to train the three networks and the mean relative error of the aggregated model (see Table~\ref{tab:FourierAccuracy}). We observe that for each width, the most accurate model is the aggregate of networks activated by SiLU and the least accurate is the Tanh aggregate. We also notice that although the models of width $300$ are more accurate than those of width $192$ the training time is much greater for a very marginal gain. Our choice is then to use networks of width $192$ and hidden layers $30\to 192\to 192\to 192 \to 10$ for further tests and compare the results obtained with ReLU and SiLU as activation functions.

\begin{table}[ht]
\caption{Mean relative error (\%) by eigenvalue and training time (s) of the surrogate depending on the width of the hidden layers of the networks and their activation function.}
\label{tab:FourierAccuracy}
\begin{center}
\setlength{\tabcolsep}{5pt}
\begin{tabular}{lccc}
& ReLU & SiLU & Tanh \\
\begin{tabular}{l}
width :\\
time :\\
$\lambda_1$ :\\
$\lambda_2$ :\\
$\lambda_3$ :\\
$\lambda_4$ :\\
$\lambda_5$ :\\
$\lambda_6$ :\\
$\lambda_7$ :\\
$\lambda_8$ :\\
$\lambda_9$ :\\
$\lambda_{10}$ :
\end{tabular}
&
\begin{tabular}{ccc}
 $90$    & $192$   & $300$   \\
 $556$   & $1097$  & $1684$  \\
 $0.144$ & $0.089$ & $0.073$ \\
 $0.168$ & $0.087$ & $0.071$ \\
 $0.249$ & $0.130$ & $0.100$ \\
 $0.229$ & $0.140$ & $0.119$ \\
 $0.275$ & $0.173$ & $0.150$ \\
 $0.285$ & $0.177$ & $0.155$ \\
 $0.347$ & $0.187$ & $0.151$ \\
 $0.378$ & $0.204$ & $0.166$ \\
 $0.315$ & $0.194$ & $0.169$ \\
 $0.373$ & $0.252$ & $0.223$ \\
\end{tabular}
&
\begin{tabular}{ccc}
 $90$    & $192$   &  $300$  \\
 $542$   & $1002$  & $1716$  \\
 $0.099$ & $0.054$ & $0.044$ \\
 $0.132$ & $0.072$ & $0.054$ \\
 $0.180$ & $0.096$ & $0.075$ \\
 $0.214$ & $0.128$ & $0.104$ \\
 $0.262$ & $0.154$ & $0.121$ \\
 $0.225$ & $0.147$ & $0.118$ \\
 $0.277$ & $0.172$ & $0.142$ \\
 $0.314$ & $0.189$ & $0.153$ \\
 $0.276$ & $0.184$ & $0.157$ \\
 $0.340$ & $0.237$ & $0.198$ \\

\end{tabular}
& 
\begin{tabular}{ccc}
 $90$    & $192$   & $300$   \\
 $842$   & $997$   & $1795$  \\
 $0.147$ & $0.097$ & $0.086$ \\
 $0.217$ & $0.112$ & $0.089$ \\
 $0.301$ & $0.170$ & $0.134$ \\
 $0.277$ & $0.188$ & $0.159$ \\
 $0.325$ & $0.223$ & $0.182$ \\
 $0.328$ & $0.221$ & $0.184$ \\
 $0.398$ & $0.244$ & $0.189$ \\
 $0.429$ & $0.264$ & $0.200$ \\
 $0.371$ & $0.242$ & $0.194$ \\
 $0.472$ & $0.308$ & $0.256$ \\
\end{tabular}
\end{tabular}
\end{center}
\end{table}

\begin{figure}[ht]
\centering
\begin{minipage}{0.14\linewidth}
\centering
\includegraphics[width=\linewidth]{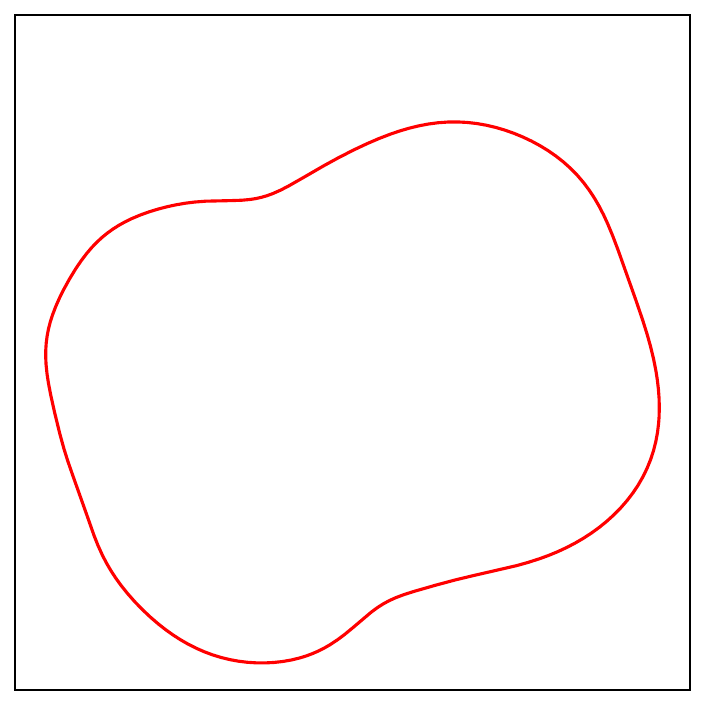}\\[-1pt]
\footnotesize $\lambda_5 = 79.72$
\end{minipage}\hfill
\begin{minipage}{0.14\linewidth}
\centering
\includegraphics[width=\linewidth]{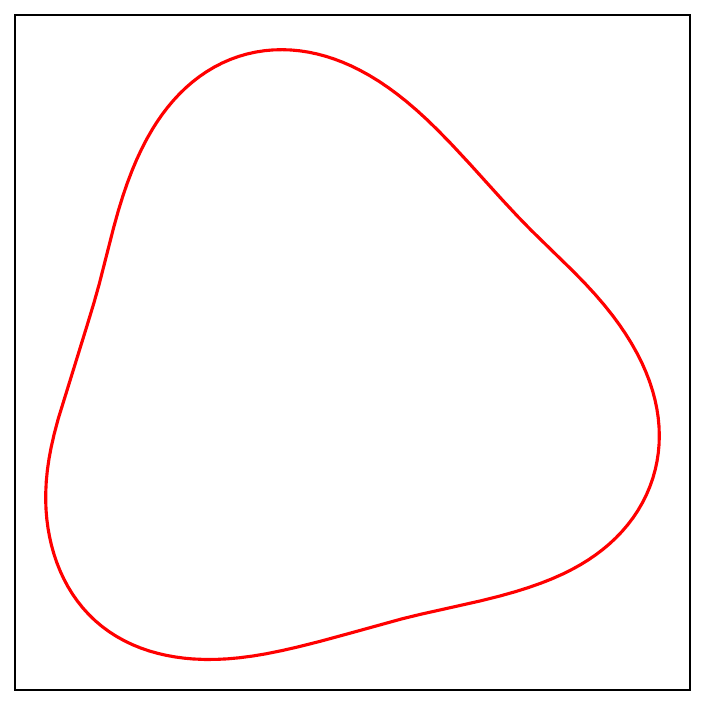}\\[-1pt]
\footnotesize $\lambda_6 = 89.21$
\end{minipage}\hfill
\begin{minipage}{0.14\linewidth}
\centering
\includegraphics[width=\linewidth]{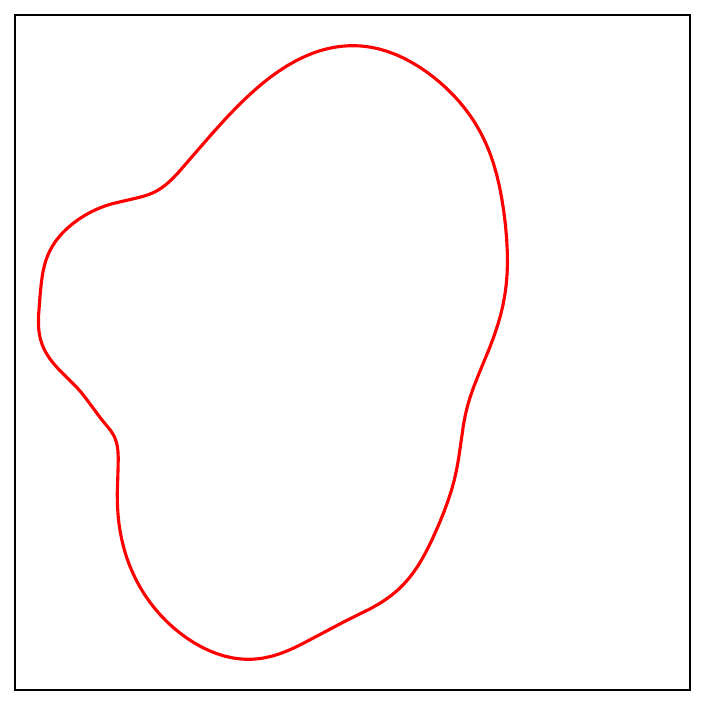}\\[-1pt]
\footnotesize $\lambda_7 = 109.17$
\end{minipage}\hfill
\begin{minipage}{0.14\linewidth}
\centering
\includegraphics[width=\linewidth]{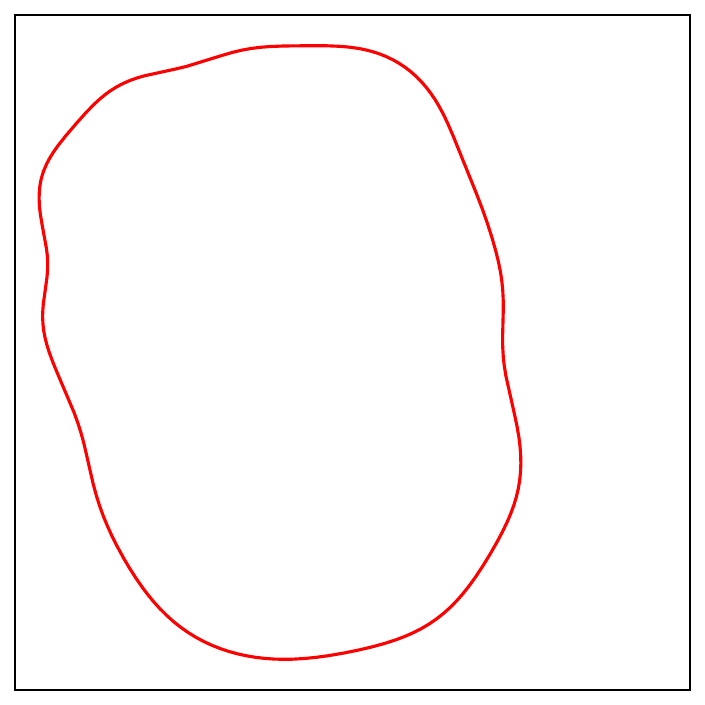}\\[-1pt]
\footnotesize $\lambda_8 = 120.83$
\end{minipage}\hfill
\begin{minipage}{0.14\linewidth}
\centering
\includegraphics[width=\linewidth]{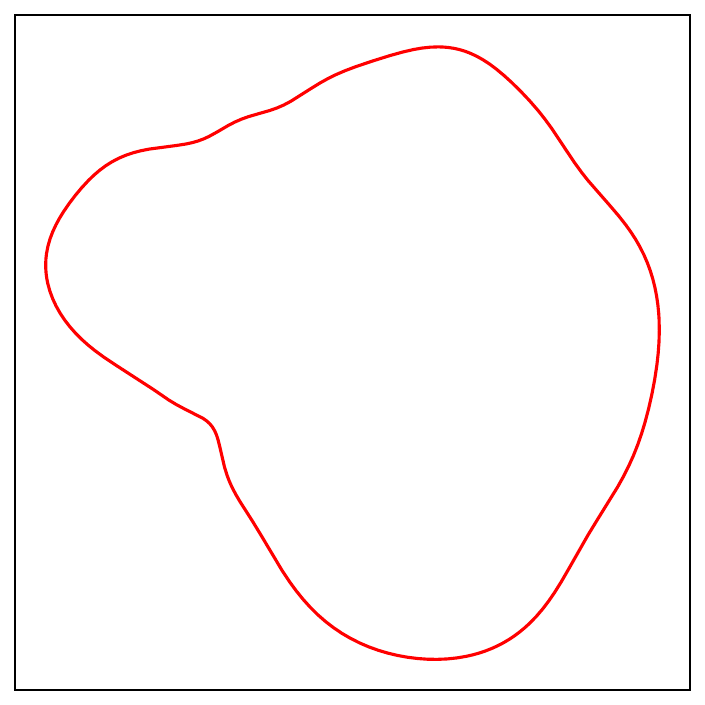}\\[-1pt]
\footnotesize $\lambda_9 = 137.13$
\end{minipage}\hfill
\begin{minipage}{0.14\linewidth}
\centering
\includegraphics[width=\linewidth]{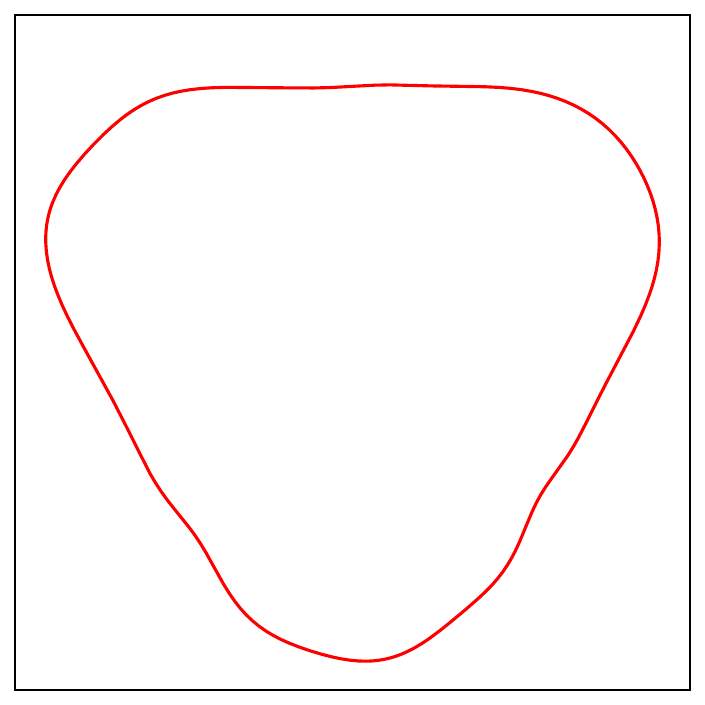}\\[-1pt]
\footnotesize $\lambda_{10} = 144.31$
\end{minipage}\hfill
\\[2pt]
\begin{minipage}{0.14\linewidth}
\centering
\includegraphics[width=\linewidth]{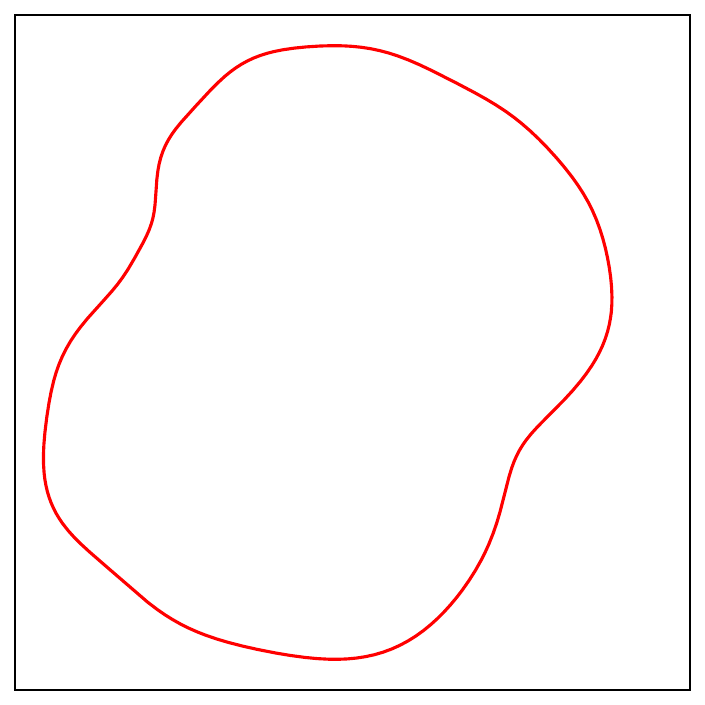}\\[-1pt]
\footnotesize $\lambda_5 = 79.96$
\end{minipage}\hfill
\begin{minipage}{0.14\linewidth}
\centering
\includegraphics[width=\linewidth]{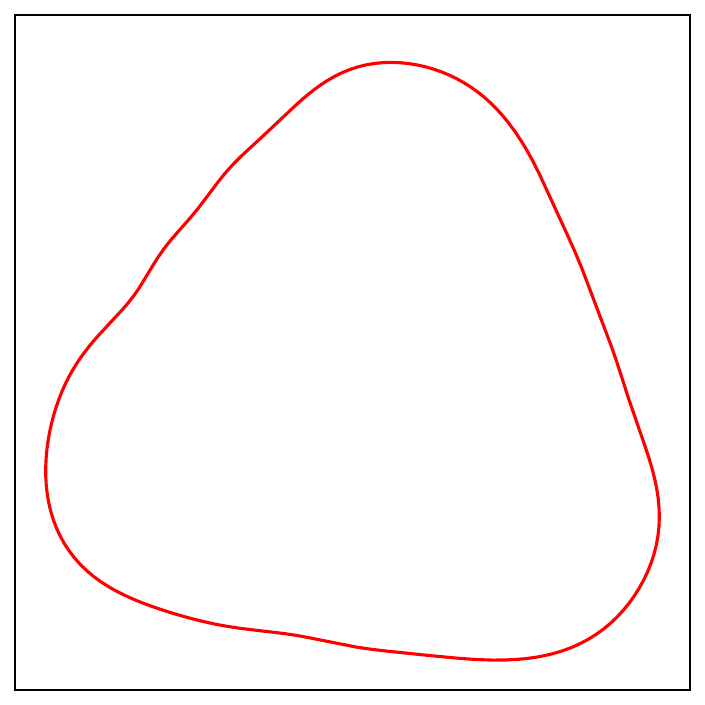}\\[-1pt]
\footnotesize $\lambda_6 = 89.24$
\end{minipage}\hfill
\begin{minipage}{0.14\linewidth}
\centering
\includegraphics[width=\linewidth]{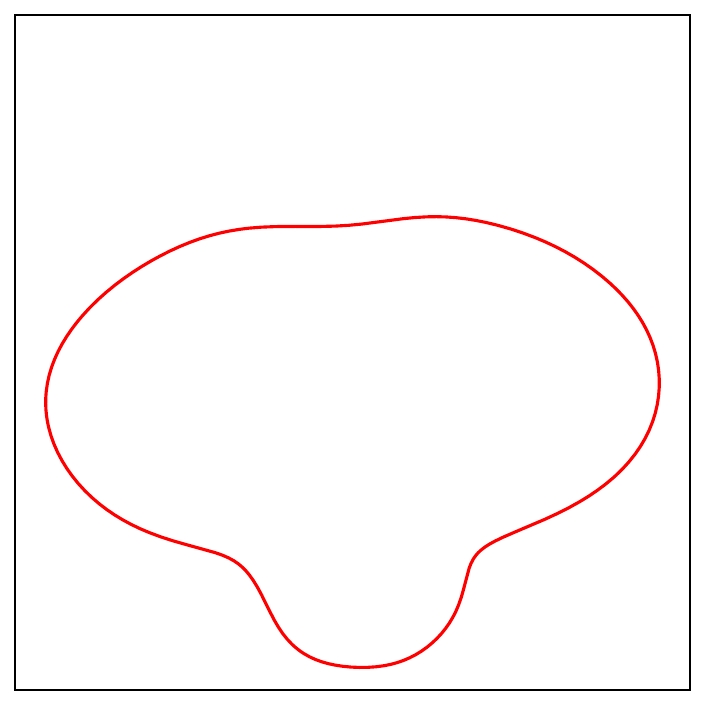}\\[-1pt]
\footnotesize $\lambda_7 = 108.89$
\end{minipage}\hfill
\begin{minipage}{0.14\linewidth}
\centering
\includegraphics[width=\linewidth]{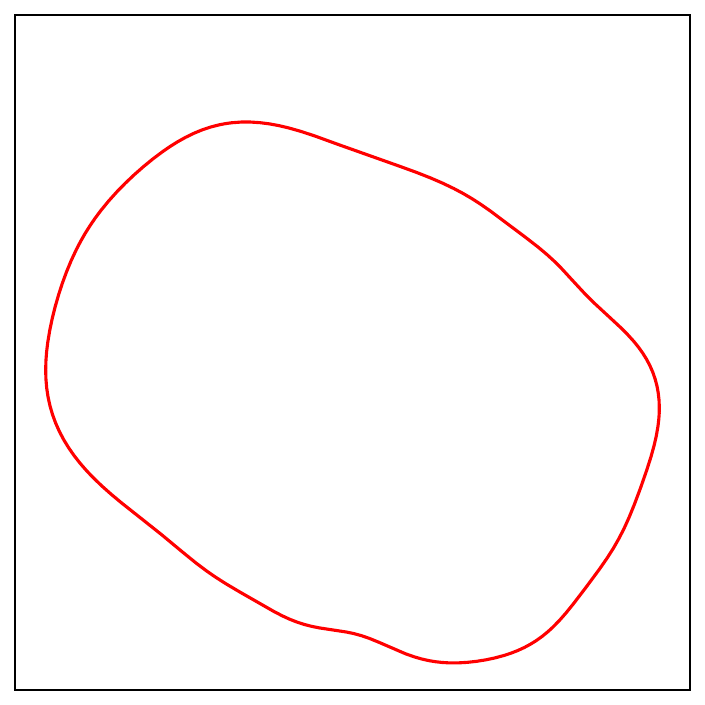}\\[-1pt]
\footnotesize $\lambda_8 = 120.77$
\end{minipage}\hfill
\begin{minipage}{0.14\linewidth}
\centering
\includegraphics[width=\linewidth]{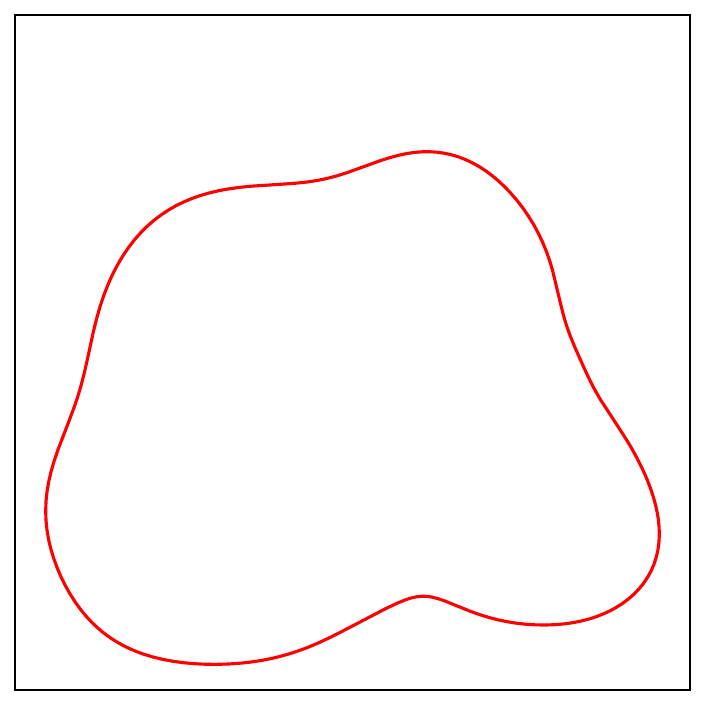}\\[-1pt]
\footnotesize $\lambda_9 = 138.69$
\end{minipage}\hfill
\begin{minipage}{0.14\linewidth}
\centering
\includegraphics[width=\linewidth]{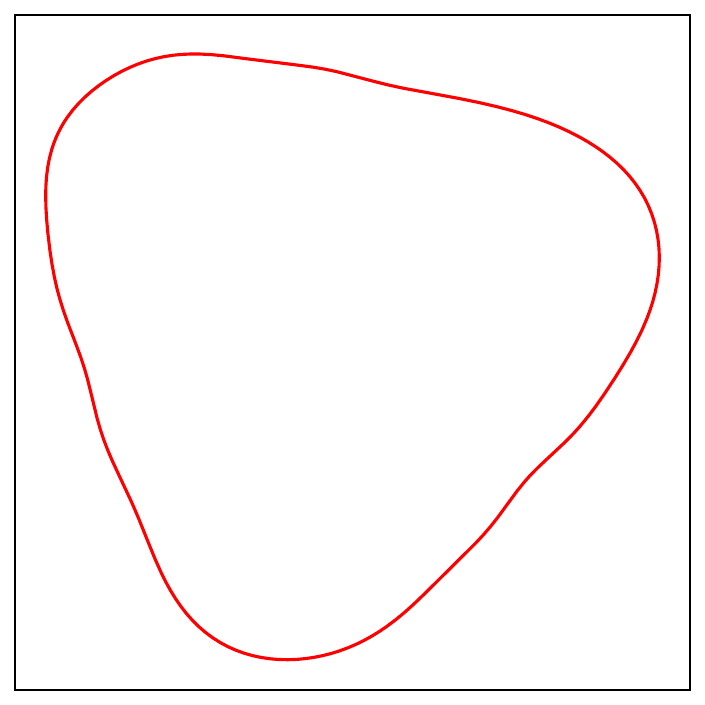}\\[-1pt]
\footnotesize $\lambda_{10} = 144.08$
\end{minipage}\hfill
\caption{Optimized shapes for $\{\lambda_5, \dots,\lambda_{10}\}$ with activation function ReLU (top) and SiLU (bottom) and associated eigenvalues computed with the PDE solver.}
\label{fig:optimReLUSiLU}
\end{figure}

Now performing tests to verify that it is possible to use our models to perform optimization we choose to compare their performance while optimizing $\lambda_5$ through $\lambda_{10}$ among sets of fixed area. During the optimization process, we follow the partial derivatives of the surrogate obtained through an automatic differentiation of the composing networks and we couple this gradient descent with a projection of the new shape on $\mathcal U_{15}$. We initialize the optimization on ten different shapes: the disk, the four best shapes of the training data for the given eigenvalue and five randomly chosen shapes among the rest of training shapes. We then keep the best of the resulting optimized shapes. We use the same starting shapes for the two models activated by ReLU and SiLU and compare their outputs by computing the eigenvalues of the two resulting optimized shapes with the FreeFEM solver (Figure~\ref{fig:optimReLUSiLU}). We observe that the results are similar for both activation function and that there is no clear better function. Both models give shapes that are close to those of Antunes and Freitas in aspect and in eigenvalue and would be good candidates to initialize an optimization process involving a traditional PDE solver. 

\begin{figure}[ht]
\centering
\begin{minipage}{0.25\linewidth}
\centering
\includegraphics[width=\linewidth]{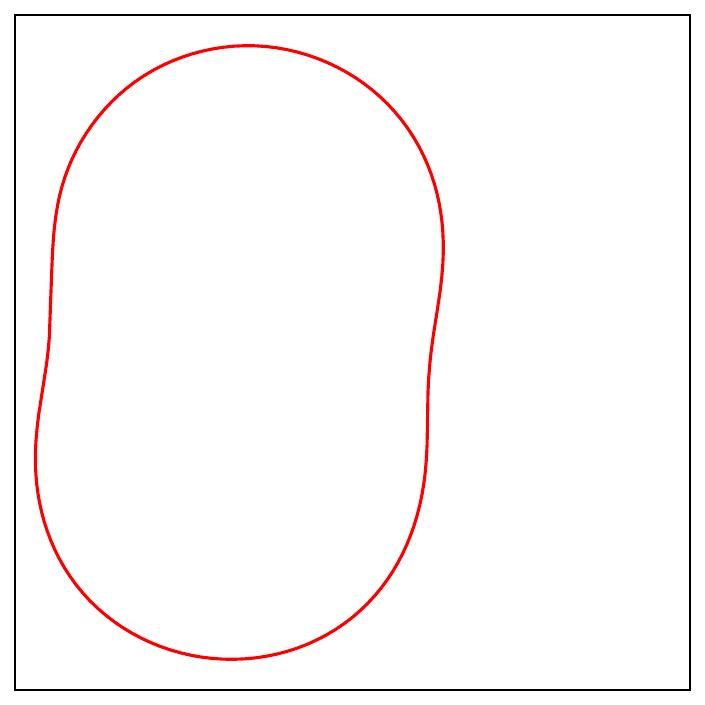}\\[-1pt]
\footnotesize $\min \lambda_1 + \lambda_2 $
\end{minipage}\hfill
\begin{minipage}{0.25\linewidth}
\centering
\includegraphics[width=\linewidth]{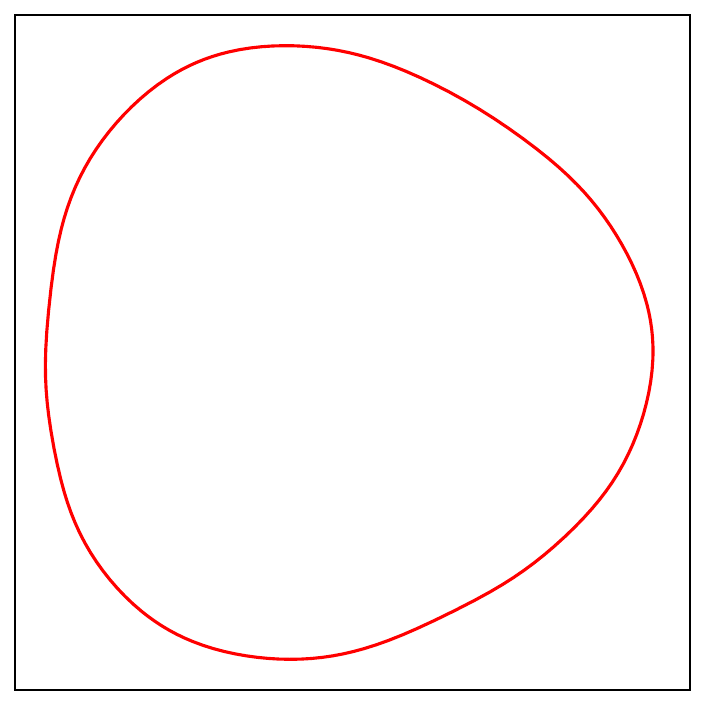}\\[-1pt]
\footnotesize $\min \sum_{i = 1}^{10} \lambda_i $
\end{minipage}\hfill
\begin{minipage}{0.25\linewidth}
\centering
\includegraphics[width=\linewidth]{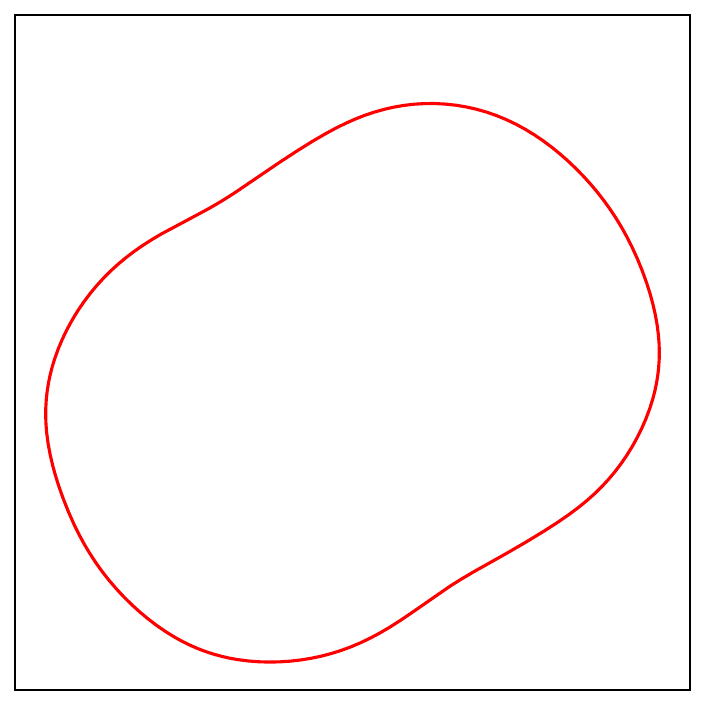}\\[-1pt]
\footnotesize $\max \lambda_6 / \lambda_1$ 
\end{minipage}\hfill
\begin{minipage}{0.25\linewidth}
\centering
\includegraphics[width=\linewidth]{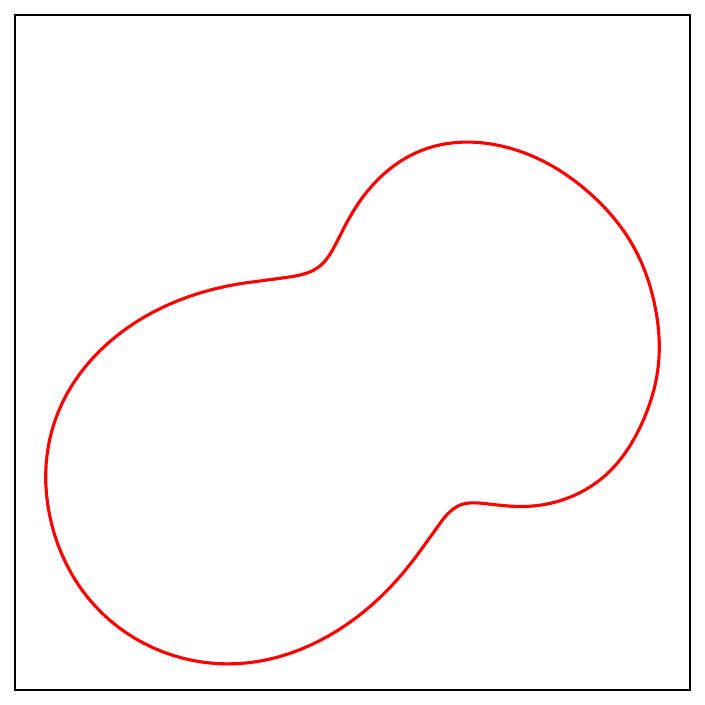}\\[-1pt]
\footnotesize $\min \lambda_2 + \cos(\lambda_5 - \lambda_1) $ 
\end{minipage}\hfill
\caption{Optimal shapes for different functions $F$.}
\label{fig:FourierOptimExamples}
\end{figure}

We therefore retain the function SiLU as it gives a slightly better accuracy and since it is smooth, so is the resulting model. Having a smooth model would allow for more involved optimization processes involving for instance second order derivatives. With this retained model coupled with the library SymPy we are able to perform optimization of any $C^1$ function of the ten first eigenvalues among sets in $\mathcal U_{15}$ of area $1$. We give in Figure~\ref{fig:FourierOptimExamples} some examples, note all the proposed functionals indeed admit optimizers as we work in the class of simply connected shapes.

\section{Topological aware model with the landscape function }

\subsection{Approach}
In this section, we consider as input representation the landscape function $w_\Omega$, introduced in Section~\ref{sec:Background}, together with the indicator function
\begin{equation*}
\mathbf{1}_\Omega(x) = 1 \text{ if } x\in\Omega\cup\partial\Omega, \qquad \mathbf{1}_\Omega(x) = 0 \text{ otherwise.}
\end{equation*}
In contrast to the SDF, the variations of the landscape function $w_\Omega$ are consistent with variations of the Dirichlet spectrum, both are robust to capacity-vanishing perturbations, as discussed in Section~\ref{sec:Background}.

Since $\mathbf 1_\Omega$ is not differentiable, we replace it by a smoothed $W_\Omega = \sigma(k\, w_\Omega / w_{\max})$, where $\sigma$ is the sigmoid function. The network input is a 3-channel image on a $64\times64$ grid. The first channel is the smoothed indicator $W_\Omega$, the second one is the normalized landscape $w_\Omega / w_{\max} \in [0,1]$, and the third one is the squared gradient $|\nabla w_\Omega / w_{\max}|^2$.

Contrary to the Fourier coefficients of Section~\ref{sec:fourier}, our grid image has no rotation equivariance/invariance: the discretization and the resampling to $64\times64$ all yields non rotation invariance. We thus rely on data augmentation during training, we simply apply rotations by $\frac{\pi}{2}$ together with a horizontal flip.

We use a U-Net encoder and decoder with skip connections, groupNorm and GELU activations. The decoder outputs $K=10$ candidate functions on the grid. The output is orthonormalized thanks to a differentiable Gram-Schmidt process, this leads to eigenfunctions approximations $\hat u_1,\dots,\hat u_K$. The ratios $\hat\lambda_k/\hat\lambda_1$, $k=2,\dots,K$, are computed with the Rayleigh quotient of each $\hat u_k$. A scalar head applied to the pooled encoder bottleneck predicts $\log(\hat\lambda_1 \cdot w_{\max})$.  We compare different architectures with the same representation and eigenvalues in the following subsection.

The loss during training is the sum of three terms. More precisely, a $L^1$ loss on the ratios $\hat\lambda_k/\hat\lambda_1$, an $L^1$ loss on $\log(\hat\lambda_1\cdot w_{\max})$, a Rayleigh-quotient term \cite{kharazmi2019variational} on $\hat u_k$ directly.

\begin{figure}[ht]
\centering
\begin{minipage}{0.135\linewidth}
\centering
\includegraphics[width=\linewidth]{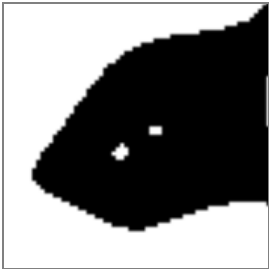}\\[-1pt]
\footnotesize holed
\end{minipage}\hfill
\begin{minipage}{0.135\linewidth}
\centering
\includegraphics[width=\linewidth]{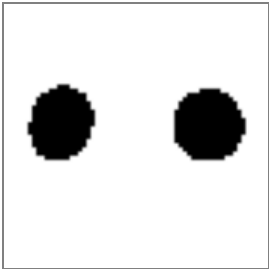}\\[-1pt]
\footnotesize several parts
\end{minipage}\hfill
\begin{minipage}{0.135\linewidth}
\centering
\includegraphics[width=\linewidth]{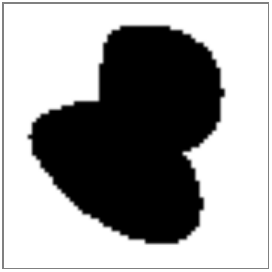}\\[-1pt]
\footnotesize nonconvex
\end{minipage}\hfill
\begin{minipage}{0.135\linewidth}
\centering
\includegraphics[width=\linewidth]{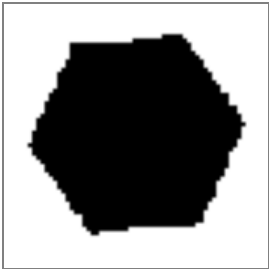}\\[-1pt]
\footnotesize polygon
\end{minipage}\hfill
\begin{minipage}{0.135\linewidth}
\centering
\includegraphics[width=\linewidth]{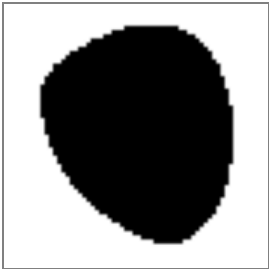}\\[-1pt]
\footnotesize simple
\end{minipage}\hfill
\begin{minipage}{0.135\linewidth}
\centering
\includegraphics[width=\linewidth]{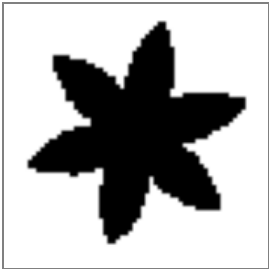}\\[-1pt]
\footnotesize star
\end{minipage}\hfill
\begin{minipage}{0.135\linewidth}\centering
\includegraphics[width=\linewidth]{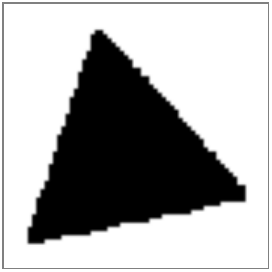}\\[-1pt]
\footnotesize triangle
\end{minipage}
\caption{One representative domain for each different shape.}
\label{fig:dataset_shapes}
\end{figure}

Concerning the dataset, we generate $N=100000$ domains including near-circular, star-shaped, non-convex, multiply-connected, several parts, polygons and equilateral triangles. For some of the domains, we compute $\Phi = h - r^2/4$, where $h$ is harmonic ($\Delta h=0$) and $\Delta(r^2/4)=1$, this gives $w_\Omega=\Phi|_\Omega$ analytically, with no expensive solving. For polygons, $w_\Omega$ is obtained by finite differences on a subset of a $\mathbb{Z}^2$ grid. Note that this does not lead to large errors since the following subsection shows that polygons are the shapes associated to the smallest relative errors. Across the dataset, $\lambda_1$ ranges over $[1.88, 751.89]$. 
Furthermore, the input domains are approximated by a discretisation which induces some errors on the prediction of the spectrum. On a grid, a finite difference used to compute the eigenvalues only needs the discrete boundary of the domain whereas a FEM method constructs a mesh that approximate more precisely the domain boundary. We then trained our model on FreeFEM computed eigenvalues, leading to a mean relative error of $1.15\%$ (median $0.53\%$) over all $K$ eigenvalues. Figure~\ref{fig:dataset_shapes} illustrates some representative examples of our dataset.

\subsection{Experiments and ablation study} 
We evaluate our model and method after training. Firstly, Table~\ref{tab:lambda_mean_relative_error} gives the mean relative error for each eigenvalue during the test of 15000 shapes not seen during training. 

\begin{table}[ht]
\caption{Mean relative test error for each eigenvalue after validation.}
\label{tab:lambda_mean_relative_error}
\centering
\footnotesize
\setlength{\tabcolsep}{5pt}
\begin{tabular}{lcccccccccc}
 & $\lambda_1$ & $\lambda_2$ & $\lambda_3$ & $\lambda_4$ & $\lambda_5$ & $\lambda_6$ & $\lambda_7$ & $\lambda_8$ & $\lambda_9$ & $\lambda_{10}$ \\
our model & $0.24\%$ & $1.08\%$ & $1.07\%$ & $1.07\%$ & $0.98\%$ & $1.05\%$ & $1.01\%$ & $0.99\%$ & $0.98\%$ & $0.85\%$ \\
\end{tabular}
\end{table}

The Table~\ref{tab:lambda_mean_relativeerror_shape} gives a comparison per shape type.

\begin{table}[ht]
\caption{Mean relative test error for each shape after validation.}
\label{tab:lambda_mean_relativeerror_shape}
\centering
\footnotesize
\begin{tabular}{lccccccc}
 & simple & star & polygon & triangle & non-convex & holed & several parts \\
Our model & $0.33\%$ & $1.13\%$ & $0.29\%$ & $0.47\%$ & $2.55\%$ & $2.33\%$ & $0.97\%$ \\
\end{tabular}
\end{table}

\textbf{Models comparison}
We compare our model, FNO and DeepONet after training on the same dataset of $N=100000$ domains. We compare these architectures under two input representations, first the landscape-based, and second an indicator function described previously. All models are trained for $300$ epochs, and evaluated by the mean relative error over $\{\lambda_1,\dots,\lambda_{10}\}$ on a test set.

\begin{table}[ht]
\caption{Mean relative error of $\{\lambda_1,\dots,\lambda_{10}\}$, by architecture and input representation.}
\label{tab:ablation_torsion_architecture}
\centering
\begin{tabular}{lcc}
Architecture & landscape function & Indicator \\
our model & $\mathbf{0.93\%}$ & $11.54\%$ \\
FNO      & $2.33\%$ & $14.71\%$ \\
DeepONet & $1.47\%$ & $11.57\%$ \\
\end{tabular}
\end{table}

Replacing the landscape-based representation by the indicator alone increases the error by a factor of $6$ to $10$. That shows that thanks to the landscape function, the model can recover the domain shape. Furthermore, for FNO and DeepONet, the indicator function based models show a train/validation gap happening after the first $20$ epochs making the indicator function impossible to generalize from our dataset.

If we replace the landscape function by the SDF function and keeping the same architecture and adding $\log(\text{SDF}_{\max})$ results in a test relative error of $33.06\%$. It is even worse for DeepONet model and FNO, with respectively $45\%$ and $54.68\%$

\textbf{Generalization to existing dataset (MPEG-7)}
\label{sec:mpeg7}
To show the generalization ability of our model, we evaluate our final trained model with $1400$ binary silhouettes from the MPEG-7 CE-Shape-1 dataset~\cite{855850}. This dataset includes $70$ classes from animals and tools. Ground truth is computed by finite differences on a $200\times200$ grid. The global mean error to $4.83\%$, please refer to Table~\ref{tab:mpeg7_error} for per eigenvalue error.

\begin{table}[ht]
\caption{Relative error per eigenvalue on MPEG-7, mean relative error $6.88\%$, median relative error $4.83\%$.}
\label{tab:mpeg7_error}
\centering
\footnotesize
\setlength{\tabcolsep}{3pt}
\begin{tabular}{lcccccccccc}
 & $\lambda_1$ & $\lambda_2$ & $\lambda_3$ & $\lambda_4$ & $\lambda_5$ & $\lambda_6$ & $\lambda_7$ & $\lambda_8$ & $\lambda_9$ & $\lambda_{10}$ \\
mean   & $2.9\%$  & $8.4\%$  & $8.0\%$  & $7.6\%$  & $7.2\%$  & $7.5\%$  & $6.7\%$  & $6.5\%$  & $6.7\%$  & $7.5\%$  \\
median & $1.5\%$  & $7.4\%$  & $6.9\%$  & $5.5\%$  & $5.4\%$  & $5.3\%$  & $4.6\%$  & $4.3\%$  & $4.2\%$  & $5.0\%$  \\
\end{tabular}
\end{table}

Our model performs well both on convex and non-convex domains. 

\textbf{Shape optimization case studies} In a similar fashion as the previous section, we experiment shape optimization by gradient descent through the surrogate and check the result against two classical spectral optimization theorems (Faber-Krahn and Hong-Krahn-Szego) and one example from the previous section.

For $\lambda_1$, we start from an ellipse of aspect ratio $a/b=6$, the gradient descent converges to $a/b=1.02$, with $\hat\lambda_1$ decreasing from $104.96$ to $36.86$, only $1.43\%$ above the exact value $j_{01}^2\pi/A=36.34$. The relative error of the resulting landscape function has also a low value of $2.35\%$ compared to the theoretical landscape. In a second experiment we start from a disk of fixed radius with two circular holes of radius $r$. $\hat\lambda_1$ decreases monotonically as $r\to0$ in every one of $20$ tested $r$. .

\begin{figure}[ht]
\centering
\begin{minipage}{0.49\linewidth}
\centering
\includegraphics[width=0.65\linewidth]{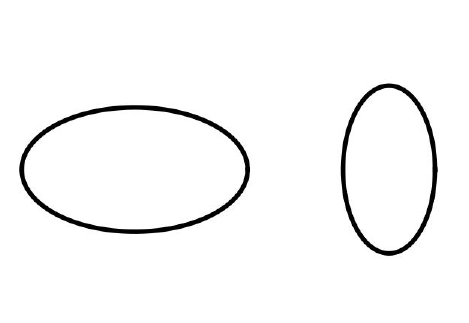}
\end{minipage}
\hfill
\begin{minipage}{0.49\linewidth}
\centering
\includegraphics[width=0.65\linewidth]{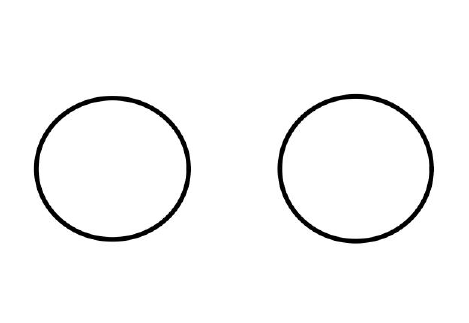}
\end{minipage}
\vspace{-2em}
\begin{minipage}{0.49\linewidth}
\centering
\includegraphics[width=0.56\linewidth]{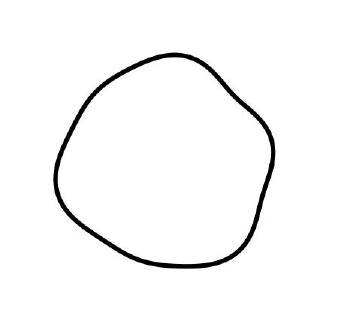}
\end{minipage}
\begin{minipage}{0.49\linewidth}
\centering
\includegraphics[width=0.56\linewidth]{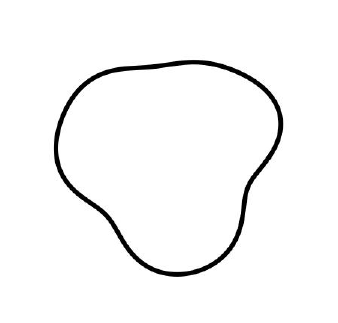}
\end{minipage}
\caption{On the left, we show the starting domain's boundary, on the right, the boundary of the domain after the gradient descent. On top, $\hat\lambda_2$ converges to the exact value for two equal disks ($2.35\%$ of relative error). After 100 iterations $a_1/b_1=1.08$, $a_2/b_2=1.05$, $R_1/R_2=0.99$.  On bottom, $\hat\lambda_6$ does not hallucinate with respect to the eigenvalue computed with FreeFem ($2.5\%$ of relative error).}
\label{fig:optim_two_ellipses}
\end{figure}

For $\lambda_2$, it is also well known that the minimizer over domains of constant area is the union of two equal disks. We start from two eccentric ellipses with size ratio $R_1/R_2=1.35$, aspect ratios $a_1/b_1=1.82$ and $a_2/b_2=0.55$ and optimize their three shape parameters (relative size, both eccentricities) to minimize $\hat\lambda_2$. The gradient descent converges to $R_1/R_2=0.99$, $a_1/b_1=1.08$, $a_2/b_2=1.05$, while $\hat\lambda_2$ decreases from $111.52$ to $74.38$, only $2.35\%$ above the exact theoretical value $2\pi j_{01}^2/A=72.67$, please refer to Figure~\ref{fig:optim_two_ellipses}. Figure~\ref{fig:optim_two_ellipses} also shows the minimizer found for $\lambda_6$ presented by Antunes and Freitas. We start from a near circle and it converges to a rounded triangle. The resulting predicted eigenvalue has a $2.5$ relative error compared to the eigenvalue computed with FreeFem. 

\section{Conclusion and perspectives}
In this study we proposed two very different approaches to tackle the problem of spectral optimization. They differ in the methods used as well as in their scopes. 

The first approach, the Fourier based model is purposely build to answer the needs of mathematicians looking to support or disprove a given conjecture. As such, the aim here is to obtain a very light surrogate to the PDE solver that may be used alongside the PDE solver and not just in its stead. The design we propose is then constrained to use a shape formalism that is also efficient for traditional numerical shape optimization and this limits the possibility for the model to be generalized to larger classes of shapes. The natural extension of this approach would be to learn other quantities such as the torsional rigidity, the Neumann eigenvalues or geometric quantities.

The second approach aims to answer the question on a more fundamental level up to loosing some precision. It uses the landscape function to represent the shapes which is the classical tool used in the analytical aspect of the field, for instance through the theory of $\gamma$-convergence. It showcases that this function is indeed the right notion of indicator function for sets when dealing with quantities linked to the Dirichlet laplacian. It would be of interest to see if, using the same procedure, it is also possible to learn the spectrum of capacitary measures which are the natural relaxation of sets in this context and can also be represented with a notion of torsion function.

\subsection*{AI use statement}
In this work, we used generative AI tools to expand our first codes as we explored new ideas. 
We have not used generative AI tools for any other purposes such as drafting the present paper or proving mathematical claims. 
We have reviewed all AI assisted work. All AI produced code has been checked by at least two authors and most of it has been manually rewritten.
We take responsibility for the final content of this work, including text, claims or artifacts produced with the aid of generative AI.
	
\subsection*{Reproducibility statement}\label{sec:reproducibility}
For reproducibility purposes, all random parameters and shapes have been generated with a fixed seed and every experiment can be reproduced.

The codes are available on a github repository \url{https://github.com/devillerochealexis/Neural-networks-for-spectral-optimization}.

The weights of the trained torsion function model are available at \url{https://doi.org/10.5281/zenodo.22943009}.

We also designed a web app to bundle the Fourier based model, it is available at \url{https://devillerochealexis.github.io/Neural-networks-for-spectral-optimization/}
 
\printbibliography

@article{AF12,
 author = {Antunes, Pedro R. S. and Freitas, Pedro},
 title = {Numerical optimization of low eigenvalues of the {Dirichlet} and {Neumann} laplacians},
 fjournal = {Journal of Optimization Theory and Applications},
 journal = {J. Optim. Theory Appl.},
 volume = {154},
 number = {1},
 pages = {235--257},
 year = {2012},
 doi = {10.1007/s10957-011-9983-3}
}

@article{AFST_2025_6_34_2_315_0,
     author = {Guy David and Antoine Gloria and Svitlana Mayboroda},
     title = {The landscape function on {$\mathbb{R}^d$}},
     journal = {Annales de la Facult{\'e} des sciences de Toulouse : Math{\'e}matiques},
     pages = {315--337},
     publisher = {Universit{\'e} de Toulouse, Toulouse},
     volume = {Ser. 6, 34},
     number = {2},
     year = {2025},
     doi = {10.5802/afst.1814},
     language = {en},
     url = {https://afst.centre-mersenne.org/articles/10.5802/afst.1814/}
}

@INPROCEEDINGS{855850,
  author={Latecki, L.J. and Lakamper, R. and Eckhardt, T.},
  booktitle={Proceedings IEEE Conference on Computer Vision and Pattern Recognition. CVPR 2000 (Cat. No.PR00662)}, 
  title={Shape descriptors for non-rigid shapes with a single closed contour}, 
  year={2000},
  volume={1},
  number={},
  pages={424-429 vol.1},
  doi={10.1109/CVPR.2000.855850}}

@article{li2020fourier,
  title={Fourier neural operator for parametric partial differential equations},
  author={Li, Zongyi and Kovachki, Nikola and Azizzadenesheli, Kamyar and Liu, Burigede and Bhattacharya, Kaushik and Stuart, Andrew and Anandkumar, Anima},
  journal={arXiv preprint arXiv:2010.08895},
  year={2020}
}

@article{lu2021deepxde,
  title={DeepXDE: A deep learning library for solving differential equations},
  author={Lu, Lu and Meng, Xuhui and Mao, Zhiping and Karniadakis, George Em},
  journal={SIAM review},
  volume={63},
  number={1},
  pages={208--228},
  year={2021},
  publisher={SIAM}
}

@article{jin2021nsfnets,
  title={NSFnets (Navier-Stokes flow nets): Physics-informed neural networks for the incompressible Navier-Stokes equations},
  author={Jin, Xiaowei and Cai, Shengze and Li, Hui and Karniadakis, George Em},
  journal={Journal of Computational Physics},
  volume={426},
  pages={109951},
  year={2021},
  publisher={Elsevier}
}

@article{kharazmi2019variational,
  title={Variational physics-informed neural networks for solving partial differential equations},
  author={Kharazmi, Ehsan and Zhang, Zhongqiang and Karniadakis, George Em},
  journal={arXiv preprint arXiv:1912.00873},
  year={2019}
}

@article{lu2021learning,
  title={Learning nonlinear operators via DeepONet based on the universal approximation theorem of operators},
  author={Lu, Lu and Jin, Pengzhan and Pang, Guofei and Zhang, Zhongqiang and Karniadakis, George Em},
  journal={Nature machine intelligence},
  volume={3},
  number={3},
  pages={218--229},
  year={2021},
  publisher={Nature Publishing Group UK London}
}

@article{hecht2012new,
  title={New development in FreeFem++},
  author={Hecht, Fr{\'e}d{\'e}ric},
  journal={Journal of numerical mathematics},
  volume={20},
  number={3-4},
  pages={1--14},
  year={2012}
}

@inproceedings{ichimaru2025neural,
  title={Neural sdf for shadow-aware unsupervised structured light},
  author={Ichimaru, Kazuto and Thomas, Diego and Iwaguchi, Takafumi and Kawasaki, Hiroshi},
  booktitle={2025 IEEE/CVF Winter Conference on Applications of Computer Vision (WACV)},
  pages={287--296},
  year={2025},
  organization={IEEE}
}

@article{costabal2024delta,
  title={$\Delta$-PINNs: Physics-informed neural networks on complex geometries},
  author={Costabal, Francisco Sahli and Pezzuto, Simone and Perdikaris, Paris},
  journal={Engineering Applications of Artificial Intelligence},
  volume={127},
  pages={107324},
  year={2024},
  publisher={Elsevier}
}

@article{li2026finite,
  title={Finite Element Eigenfunction Network (FEENet): A Hybrid Framework for Solving PDEs on Complex Geometries},
  author={Li, Shiyuan and Salahshoor, Hossein},
  journal={arXiv preprint arXiv:2602.00870},
  year={2026}
}

@inproceedings{williamson2025neural,
  title={Neural geometry processing via spherical neural surfaces},
  author={Williamson, Romy and Mitra, Niloy J},
  booktitle={Computer Graphics Forum},
  volume={44},
  number={2},
  pages={e70021},
  year={2025},
  organization={Wiley Online Library}
}

@article{rowan2025solving,
  title={Solving engineering eigenvalue problems with neural networks using the Rayleigh quotient},
  author={Rowan, Conor and Evans, John and Maute, Kurt and Doostan, Alireza},
  journal={arXiv preprint arXiv:2506.04375},
  year={2025}
}

@inproceedings{park2019deepsdf,
  title={Deepsdf: Learning continuous signed distance functions for shape representation},
  author={Park, Jeong Joon and Florence, Peter and Straub, Julian and Newcombe, Richard and Lovegrove, Steven},
  booktitle={Proceedings of the IEEE/CVF conference on computer vision and pattern recognition},
  pages={165--174},
  year={2019}
}

@article{belieres2025volume,
  title={Volume-Preserving Geometric Shape Optimization of the Dirichlet Energy Using Variational Neural Networks},
  author={B{\'e}li{\`e}res, Amaury and Franck, Emmanuel and Michel-Dansac, Victor and Privat, Yannick and others},
  journal={Neural Networks},
  volume={184},
  pages={106957},
  year={2025}
}
\end{document}